\documentclass{article}

\usepackage{arxiv}

\usepackage[utf8]{inputenc} % allow utf-8 input
\usepackage[T1]{fontenc}    % use 8-bit T1 fonts
\usepackage{hyperref}       % hyperlinks
\usepackage{url}            % simple URL typesetting
\usepackage{amsfonts}       % blackboard math symbols
\usepackage{nicefrac}       % compact symbols for 1/2, etc.
\usepackage{microtype}      % microtypography
\usepackage{lipsum}
\usepackage{amsmath,amssymb,amsfonts}
\usepackage{algorithm} % For the algorithm environment
\usepackage{algpseudocode} % For the algorithmic environment
\usepackage{multirow}

\usepackage{graphicx}
\usepackage{caption}
\usepackage{subcaption}
\usepackage{booktabs}
\usepackage{enumitem}

\title{K-Means Clustering With Incomplete Data with
the Use of Mahalanobis Distances
}

\author{
 Lovis Kwasi Armah \\
  Department of Statistics, University of Minnesota\\
  Duluth, United States of America\\
  \texttt{armah007@d.umn.edu} \\
   \And
 Igor Melnykov \\
  Department of Statistics, University of Minnesota\\
  Duluth, United States of America\\
}

\begin{document}
\maketitle
\begin{abstract}
Effectively applying the K-means algorithm to clustering tasks with incomplete features remains an important research area due to its impact on real-world applications. Recent work has shown that unifying K-means clustering and imputation into one single objective function and solving the resultant optimization yield superior results compared to handling imputation and clustering separately.

In this work, we extend this approach by developing a unified K-means algorithm that incorporates Mahalanobis distances, instead of the traditional Euclidean distances, which previous research has shown to perform better for clusters with elliptical shapes.

We conducted extensive experiments on synthetic datasets containing up to ten elliptical clusters, as well as the IRIS dataset. Using the Adjusted Rand Index (ARI) and Normalized Mutual Information (NMI), we demonstrate that our algorithm consistently outperforms both standalone imputation followed by K-means (using either Mahalanobis or Euclidean distance) and K-Means with Incomplete Data, the recent K-means algorithms that integrate imputation and clustering for handling incomplete data. These results hold across both the IRIS dataset and randomly generated data with elliptical clusters. 
\end{abstract}

% keywords can be removed

\section{Introduction}
Clustering algorithms devise a systematic process to group data into similar categories using some measure. There are several different types of clustering algorithms based on how subgroups are found in data. Common types are Hierarchical Clustering, Partition Clustering, Grid based Clustering, Density based Clustering, and Soft clustering. Grid based clustering carries out clustering of high dimensional data using group of cells that form a grid structure. Common examples are Statistical information grid-based (STING) algorithm and Optimal grid (OPTIGRID) \cite{wang1997sting, hinneburg1999optimal}. Density based clustering separates low data concentration points from highly concentrated ones. Notable examples are DBSCAN \cite{deng2020dbscan} and DENCUE \cite{rehioui2016denclue}.  Soft clustering allows data points to belong to more than one cluster with a graduating degree of membership. Examples of soft clustering algorithms are fuzzy C-Means,\cite{bezdek1984fcm} and  Fuzzy clustering by Local Approximation of Membership (FLAME). 

Partitioning clustering, unlike other clustering types discovers subgroups using some objective function which improves the quality of the clustering. K-means is an example of a Hierarchical clustering that assigns data points into one of K-subgroups based on a distance closest to the center of the centroid.  K-Means is very easy to implement and as a result widely used in various fields, including computer vision, computer security, health, social sciences, and natural language processing. In computer vision, it is employed for object detection and image processing as well as algorithms in computer security helping detect Distributed Denial of Service (DDoS) attacks   In health, and social sciences, it is used for data summarization and segmentation, among other applications \cite{ahmed2020k,oyelade2019data}.

Despite the popularity of K-means,it is often applied to data with incomplete features. However, Data must be made complete so K-means algorithm can be applied. To address the problem of incompleteness, various heuristics and statistical methods are employed. Heuristics might involve dropping incomplete data coordinates, while statistical methods utilize available data properties to fill in the incomplete data. Common imputation techniques include zero-filling, conditional mean, median, mode, regression imputation, and K-nearest neighbors (KNN) \cite{wang2019k}. Other methods such as Expectation-Maximization (EM) algorithm, estimate the most likely values for the missing data. However, the problem is that these imputation methods are often applied separately from the K-means algorithm, which has been shown by \cite{wang2019k}, to be less effective compared to unifying the imputation and clustering into one objective function. 

Furthermore, K-means is well suited for spherical clusters, but it is often applied to real-world data where interactions between features distort the spherical or compact shape. To overcome the challenge of lack of spherical shapes associated with real-world data, \cite{melnykov2014k} demonstrated that employing the Mahalanobis distance rather than Euclidean distance typically used in K-means works well for non-spherical datasets. 

\begin{enumerate}[label=\roman*.]
    \item We extend existing K-means Clustering with Incomplete data algorithm to work with all clusters irrespective of the shape. We demonstrate that our extension performs better using extensive experiments done that we did using synthetic datasets with non-spherical shapes and one UCI benchmark dataset.  
    \item Extensive simulation study to determine how the level of incompleteness distribution impacts the performance of the proposed algorithm and suggest areas of improvement.
\end{enumerate}

\section{Related Works}
In this section, we review the literature on clustering algorithms, particularly on K-Means, We also discuss a variant of K-Means algorithm that uses Mahalanobis distances, We discuss imputation methods employed to deal with incomplete data and cover K-Means clustering with incomplete data. We provide a brief summary of relevant work and refer readers to their findings for detailed summaries.

\subsection{K-Means Algorithm}

K-means is an example of Partition Clustering that assigns data points into one of K groups, Each group has a center called centroid and points are assigned to the closest centroid after calculating the distance to each centroid. The algorithm requires the use of some measure to determine the closest distance to a cluster centroid. Furthermore, the algorithm requires the number of clusters \( K \) to be known and seeded with \( K \) values. Knowing the number of clusters helps to partition a set of \( n \) \( p \)-dimensional data points \( \mathbf{X} = \{\mathbf{x}_i\}_{i=1}^{n} \) into \( K \) clusters. After the initial seeding of values, it continuously changes the cluster configuration until the algorithm converges to a stable solution by reassigning points to clusters that minimize the objective function below. 

\begin{equation}
\text{obj} = \sum_{i=1}^{n} \sum_{j=1}^{k} h_{ij} d(\mathbf{x}_i, \boldsymbol{\mu}_j)),
\end{equation}
where \( h_{ij} \) is an indicator variable that is 1 if data point \( \mathbf{x}_i \) belongs to cluster \( j \) and 0 otherwise, \( \boldsymbol{\mu}_j \) is the centroid of cluster \( j \), and \( d(\mathbf{x}_i, \boldsymbol{\mu}_j) \) is the distance metric used,

The most commonly used distance metric is the Euclidean distance, although other metrics, such as the Mahalanobis distance, can also be employed. The simplicity and ease of computation make the Euclidean distance a very popular choice for the K-means algorithm. However, because the Euclidean distance does not account for the shape of the clusters in its distance calculation, it may incorrectly assign points to the closest clusters, which might differ from the assignments obtained when the cluster shape is taken into consideration. 

\subsection{K-Means with Mahalanobis distances}

K-Means with Mahalanobis distance is similar to K-means except that it replaces the distance metric with Mahalanobis distance. It deals with the problem of assigning points to wrong clusters by replacing Euclidean distance, which does not factor the shape of the clusters in the distance metric, with  Mahalanobis distance. The Mahalanobis distance takes the shape of the cluster, which is determined by the covariance matrix ${\boldsymbol{\Sigma}}$. It is defined as:

\begin{equation*}
d_M(\mathbf{x}_i, \boldsymbol{\mu}_k) = \sqrt{(\mathbf{x}_i - \boldsymbol{\mu}_k)^T {\boldsymbol{\Sigma}}_k^{-1} (\mathbf{x}_i - \boldsymbol{\mu}_k)}
\end{equation*}

where \( \boldsymbol{\mu}_k \) and ${\boldsymbol{\Sigma}_k}$ are the mean vector and covariance matrix of the $k^{th}$ cluster, \\ The use of the covariance matrix makes it less popular as it must be properly initialized to work properly. 

Melnykov et al \cite{melnykov2014k}, demonstrated that the K-Means with Mahalanobis distances algorithm consistently outperforms the traditional K-Means algorithm in clustering data with non-spherical covariance structures and that it is particularly effective in scenarios where the clusters have different sizes and orientations, which are poorly handled by the Euclidean distance metric.

\subsection{Imputation methods employed to deal with incomplete data in K-Means}
Incomplete data are a common problem associated with real-world data. Incomplete data  may arise from factors such as human factors such as forgetfulness on the part of data providers, failure of some data collecting procedure, or  dependence structure where the loss of one missing coordinates impacts another coordinates.  Missing completely at random (MCAR) is used to describe factors that result in incompleteness other than related to the data itself whereas Missing at  Random (MAR) is related to the observed coordinates in the data set. Statistical Data imputation method, such as mean and median, can be leverage for completing data  where statistical relationship exit between coordinates, Model-based imputing  methods such as regression imputation exploit the multivariate relationship between coordinates to complete data. For an imputation method to work well with K-means algorithm, the imputation must not distort the shape relationship between existing points and imputed points. In this work, We used  used mean and KNN but review below commonly used imputation methods. 

\begin{itemize}
    \item Mean imputation involves replacing missing values with the mean of the observed values for that coordinate. While this method is simple and quick, it can distort the data distribution and reduce variability \cite{little2002statistical}.

    \item Median imputation replaces missing values with the median of the observed values for that feature. This method is more robust to outliers compared to mean imputation \cite{rubin2004multiple}.
    \item  Regression imputation predicts missing values using a regression model based on other features. This method assumes a linear relationship between features \cite{little2019statistical}.
    \item Multiple imputation generates multiple imputed datasets by simulating different plausible values for the missing data. It accounts for the uncertainty of the missing values \cite{rubin2004multiple}.
    \item The Expectation-Maximization (EM) imputation iteratively estimates the missing values based on the observed data and parameters of the data distribution. It can handle complex relationships but is computationally intensive \cite{dempster1977maximum}.
    \item K-Nearest Neighbors (KNN) imputation uses the values from the K-nearest neighbors to impute the missing values. It can preserve data structure but is computationally intensive \cite{troyanskaya2001missing}.
    
\end{itemize}

However, as noted by \cite{wang2019k}, isolating the imputation method from the K-means algorithm is less effective compared to integrating the two within a unified framework.

\subsection{K-Means with Incomplete Data}

To address the problem of lack of effectiveness that occurs when the imputation method is isolated from the K-means algorithm, \cite{wang2019k} proposed a novel approach that integrates imputation and clustering to allow the K-Means algorithm to handle incomplete data effectively. The algorithm unifies the K-means clustering and imputation into one single objective function and solves the resultant optimization.  It demonstrated that their method yield superior results compared to handling imputation and clustering separately. We refer readers to Algorithm 1 of their paper for more details. 

The problem with their proposed algorithm  is that it uses euclidean distance and therefore may not work well with real world data that is non spherical.

\subsection{K-Means With Incomplete Data Using Mahalanobis distance}
\label{sec:second}

To make their proposed algorithm generalize to all real world data, we propose  the K-Means With Incomplete Data Using Mahalanobis distance. We followed the formulation proposed by \cite{wang2019k}, which dynamically fills missing values by dividing the incomplete data into observable features \( \mathbf{x}_i(o_i) \) and missing features \( \mathbf{x}_i(m_i) \). In this approach, we optimize the assignment matrix \(\mathbf{H}\), cluster centers \( \boldsymbol{\mu}_k \), missing values \( \mathbf{x}_i(m_i) \), and the covariance matrices ${\boldsymbol{\Sigma}}$, \( k = 1, 2, \dots, K \) .

Our method differs from their proposal by replacing the Euclidean distance used in updating the cluster centers \( \boldsymbol{\mu}_k \) and assignment matrix \(\mathbf{H}\) with Mahalanobis distances. Furthermore, instead of imputing the missing values \( \mathbf{x}_i(m_i) \) with their corresponding cluster centers, we use their conditional means as the algorithm progresses, thereby enhancing clustering accuracy. The details of this modified algorithm are shown in Algorithm 1.

Similar to the approach by \cite{melnykov2014k}, the solution quality is evaluated using the classification likelihood, defined as:

\[
L(\boldsymbol{\mu}, {\boldsymbol{\Sigma}}_k; k=1,2,\dots,K) = \prod_{i=1}^n \phi(x_i, \boldsymbol{\mu}, {\boldsymbol{\Sigma}}_{z_i}),
\]
where \(\phi(\cdot)\) is the pdf of the $p$-dimensional normal distribution. We then maximize \(\log L\) which is proportional to $-\sum_{k=1}^{K} n_k\sum_{i=1}^{p} \log(\hat{\lambda}_{k,i})$, where \(\hat{\lambda}_{k,i}\) are the eigenvalues of the covariance matrix ${\boldsymbol{\Sigma}}_k$ and $n_k$ is the number of points in cluster $k$. Thus, we use

\[
A = \sum_{k=1}^{K} n_k \sum_{i=1}^p \log(\hat{\lambda}_{k,i})
\]
as is the solution quality criterion. The classification that results in the highest observed value of \(A\) is reported as the best solution found.

%\textbf{Time Complexity}: Similar to the K-Means with incomplete data proposed by \cite{wang2019k}, our 
%Algorithm 1 also considers the data matrix \( X \) as another variable to 
%be optimized and therefore replaces the missing values with the conditional means of the related cluster centers. Additionally, the algorithm uses the initialization procedure proposed by \cite{melnykov2014k}, which has a time complexity of 
%\[
%O(\text{launches} \cdot \max(n^2, \text{max.it} \cdot K \cdot n \cdot \log K)),
%\]
%where:
%\begin{align*}
%n & : \text{Number of data points}, \\
%K & : \text{Number of clusters}, \\
%p & : \text{Dimensionality of the data}, \\
%\text{max.it} & : \text{Maximum number of iterations allowed}, \\
%\text{launches} & : \text{Number of runs of the algorithm}.
%\end{align*}
%For small dimensions (\( p \)), the time complexity of our algorithm simplifies to \( O(tkndy) \), where \( t \), \( k \), \( n \), and \( d \) represent the number of iterations, clusters, samples, and dimensions, respectively, and \( y \) represents the initialization procedure of the Mahalanobis procedure.

\begin{algorithm}
\caption{K-Means Clustering with the Use of Mahalanobis Distances With Incomplete Data}
\begin{algorithmic}% Enable line numbering
    \Require Incomplete data matrix $\mathbf{X} = \{\mathbf{x}_i\}_{i=1}^{n}$, number of clusters $K$, convergence tolerance $\epsilon_0$, and missing index $m$.
    \Ensure Complete data matrix $\mathbf{X}$ and assignment matrix $\mathbf{H}$.
\begin{enumerate}
    \item Input missing values
    \item Initialize and obtain a first run of the K-means algorithm using the method proposed by \cite{melnykov2014k}.
    \item \textbf{Repeat}
    \begin{itemize}
        \item Update $\mathbf{H}$ with a fixed data matrix $\mathbf{X}$, cluster centers $\boldsymbol{\mu}$, and covariance matrix $\boldsymbol{{\boldsymbol{\Sigma}}}$.
        \item Update cluster centers $\boldsymbol{\mu}$ and covariance matrix $\boldsymbol{{\boldsymbol{\Sigma}}}$.
        \item Update matrix $\mathbf{X}$ with fixed $\mathbf{H}$, cluster centers $\boldsymbol{\mu}$, and covariance matrix $\boldsymbol{{\boldsymbol{\Sigma}}}$.
    \end{itemize}
    \item \textbf{Until} $\frac{\text{obj}(t-1) - \text{obj}(t)}{\text{obj}(t)} \leq \epsilon_0$
\end{enumerate}
\end{algorithmic}
\end{algorithm}

\subsection{Experiment Settings}

\subsubsection{Datasets}
The proposed algorithm was evaluated on Iris and a generated dataset.The cluster shapes in the Iris dataset are non-spherical, as there is considerable overlap between two of the clusters. This makes the Iris dataset representative of real-world data with non-spherical cluster details.\cite{melnykov2014k}. Additionally, to evaluate the performance of the K-means algorithm on datasets with clusters exhibiting varying degrees of overlap—deviating substantially from a spherical shape—we simulated datasets using the \texttt{MIXSIM} package. \cite{melnykov2012mixsim}.\texttt{MIXSIM} allows the generation of a dataset with a specified number of clusters \( K \), number of coordinates \( P \), sample size \( n \), and a specified level of pairwise overlap, which in our case was set to \(\check{\omega} = 0.001, 0.01, 0.1\).  The generated datasets used in our experiments have 5 coordinates, 10 clusters, and 1000 records. The Iris data consists of 150 records, three (3) clusters made up of species of Iris flowers (Iris setosa, Iris versicolor, and Iris virginica), and organized into four (4) coordinates. \cite{fisher1936iris}.Table \ref{tab:dataset} shows the details of the dataset information.

\subsubsection{Experimental Method}

In our experiments, we follow the experimental design illustrated in Algorithm 2 and briefly described here. We begin with the IRIS dataset or generate a new dataset by specifying the number of coordinates \( c \), the number of clusters \( K \), and the required cluster overlap \( \check{\omega} \) using MIXSIM. Once the dataset is prepared, we introduce missing values by making one or more coordinates incomplete based on the specified percentage of missing data (\(\%d\)) ie, 10\%, 20\%, 30\%, 40\%, and 50\%. This process is repeated to generate 100 random incomplete datasets.

For each dataset, we apply an imputation method (KNN or Mean imputation) before training and evaluating each algorithm: standard K-means, K-means with incomplete data (referred to as Unified K-means), and our own proposed algorithm, referred to in the results table as K-Mahal. The performance of each algorithm is then recorded and summarized, as shown in Tables~\ref{tab:comparison}, \ref{tab:mean_imputation}, \ref{tab:nmi_comparison},  \ref{tab:comparison_generated_knn} and \ref{tab:comparison_generated_mean}.

The Adjusted Rand Index (ARI) \cite{hubert1985comparing} is used to evaluate the similarity between the output of each clustering algorithm and the true labels. Also Normalized Mutual Information (NMI) is used to  quantify the agreement between the results produced by an algorithm and the true labels \cite{strehl2002cluster}. For all algorithms, we report the median values along with the corresponding interquartile ranges. All datasets and their missing values as well as the code are implemented in R %and are available at GitHub\footnote{\url{https://github.com/your-repo-link}}.

\begin{algorithm}
\caption{Experiment Design}
\begin{algorithmic}
    \Require Data $D$, Coordinates $c$, Missing Percentage $p$, Algorithm $alg$, Runs $R$, Cluster Overlap $\check{\omega}$, Number of Clusters $k$, Imputation method $Imp$.
    
    \State $D \gets$ IRIS \; \text{or} \; \texttt{GenerateData}(c, $\acute{\omega}$, k)
    \State $IncompleteData \gets \textsc{GenerateIncompleteData}(D, c, p, 100)$
    
    \For{each $D_i \in IncompleteData$}
        \State $maxNMI, maxARI \gets (0, 0)$
        \For{$r = 1$ to $R$}
            \State $(nmi, ari) \gets \textsc{TrainAndEvaluate}(D_i, alg, Imp)$
            \State $maxNMI \gets \max(maxNMI, nmi)$
            \State $maxARI \gets \max(maxARI, ari)$
        \EndFor
        \State Append $(p, c, maxNMI, maxARI)$ to results table
    \EndFor
    
    \State \Return Performance table
\end{algorithmic}
\end{algorithm}

\begin{table}
\centering

\label{tab:dataset_missing_ratio}
\begin{tabular}{l l}
\toprule
\textbf{Dataset} & \textbf{Missing Ratio} \\
\midrule
Iris & 10\%-50\% \\
genp5k101000 & 10\%-50\% \\

\bottomrule
\end{tabular}
\hspace{1in}
\caption{Summary of data and incomplete ratios used in clustering experiments.}
  \label{tab:missingValues}
\end{table}

\hspace{1in}

\begin{table}
\centering
%\resizebox{\textwidth}{!}{%
\begin{tabular}{llll}
\toprule
\textbf{Dataset} & \textbf{\#Samples} & \textbf{\#Dimensions} & \textbf{\#Clusters} \\
\midrule
Iris         & 150       & 4            & 3          \\
genp5k101000  & 1000      & 5            & 10         \\

\bottomrule
\end{tabular}%
\hspace{1in}
%}
\caption{Summary of the number of samples, dimensions, and cluster details of the dataset used in the clustering experiments.}

\label{tab:dataset}
\end{table}

\section{Results}

\begin{figure}[hbt]
    \centering
    \begin{tabular}{cc}
        \begin{subfigure}{0.45\linewidth}
            \centering
            \includegraphics[width=\linewidth]{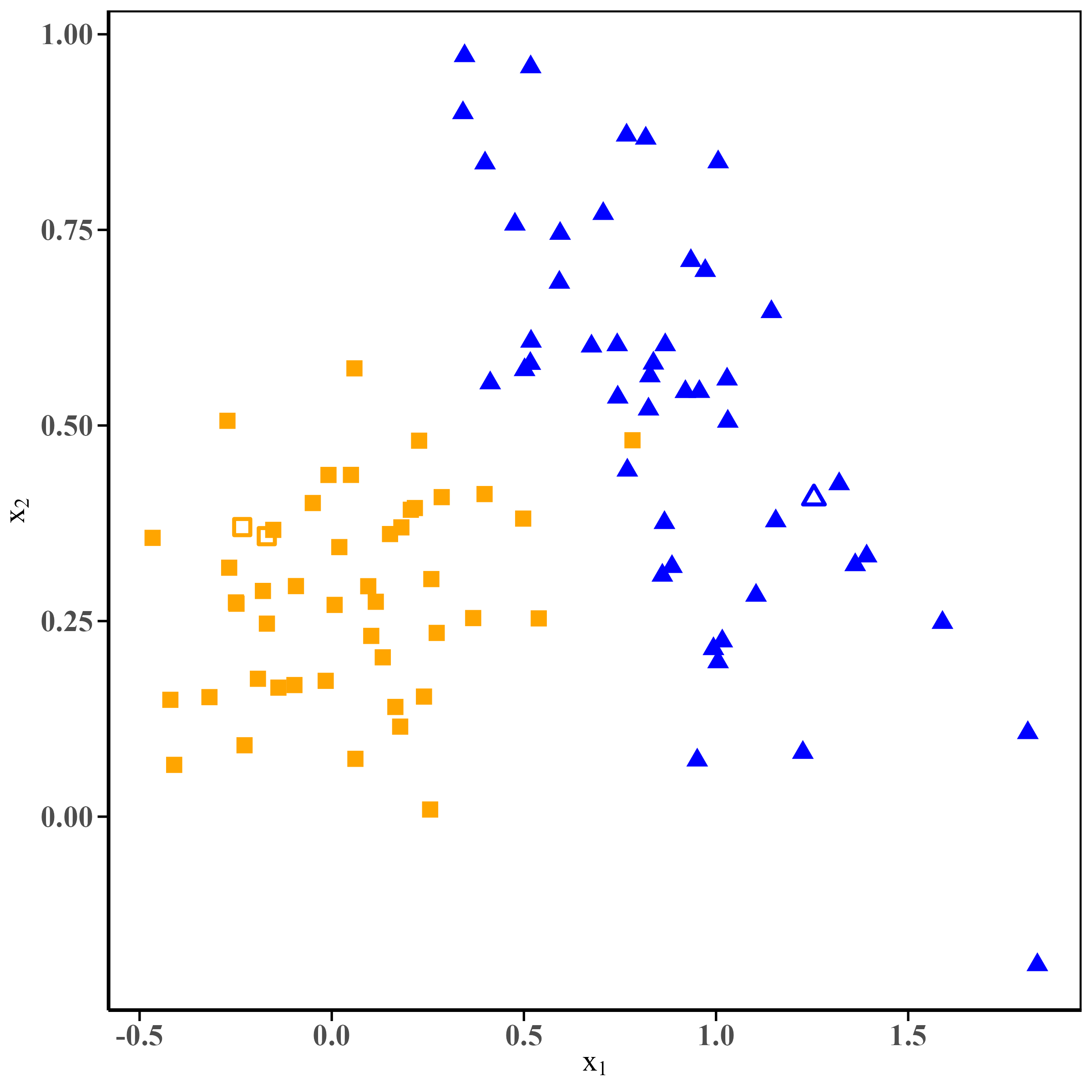}

            \caption{Cluster Distribution with Incomplete Values.}
            \label{fig:cluster1}
        \end{subfigure} &
        \begin{subfigure}{0.45\linewidth}
            \centering
            \includegraphics[width=\linewidth]{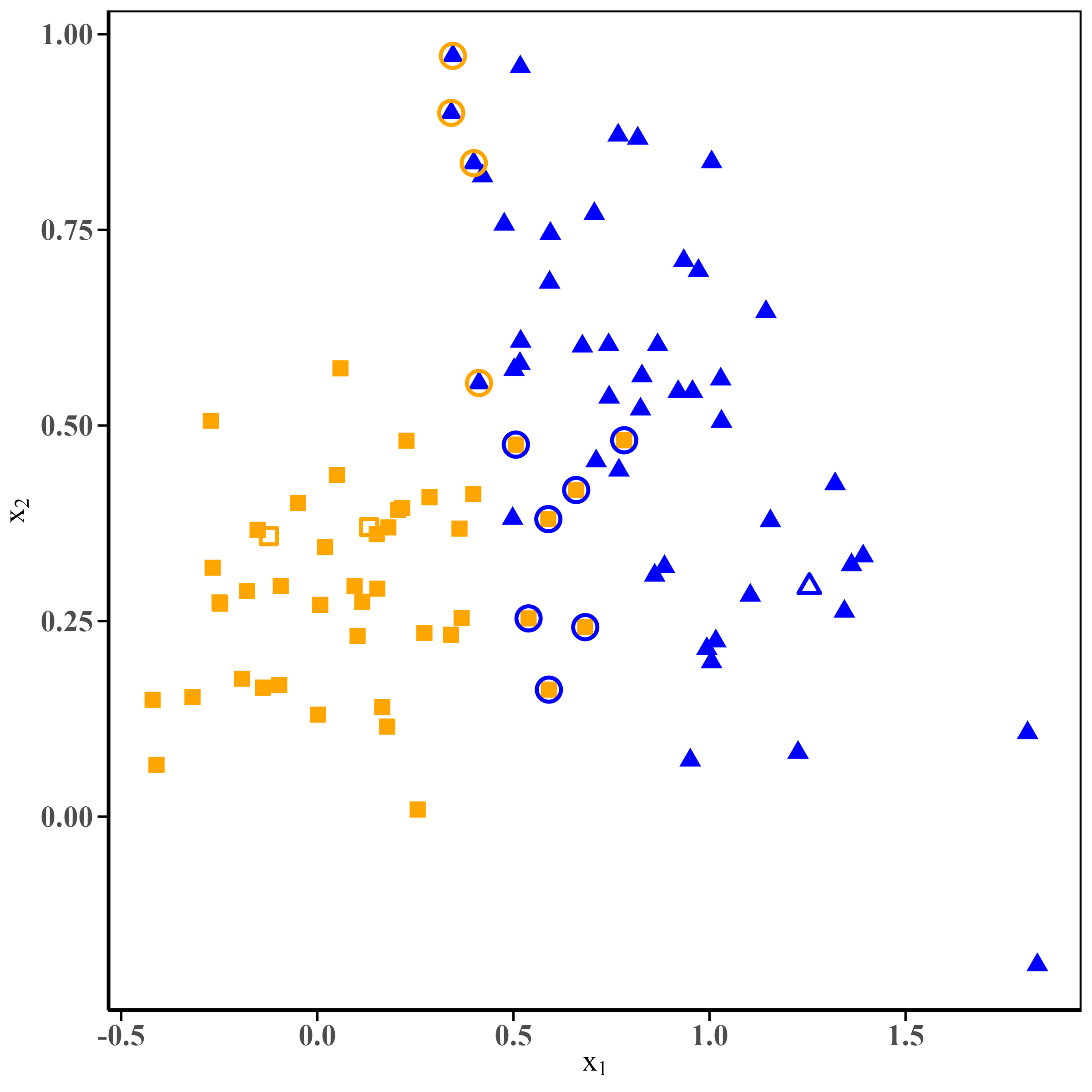}

            \caption{K-MEANS Clustering}
            \label{fig:cluster2}
        \end{subfigure} \\
        \begin{subfigure}{0.45\linewidth}
            \centering
            \includegraphics[width=\linewidth]{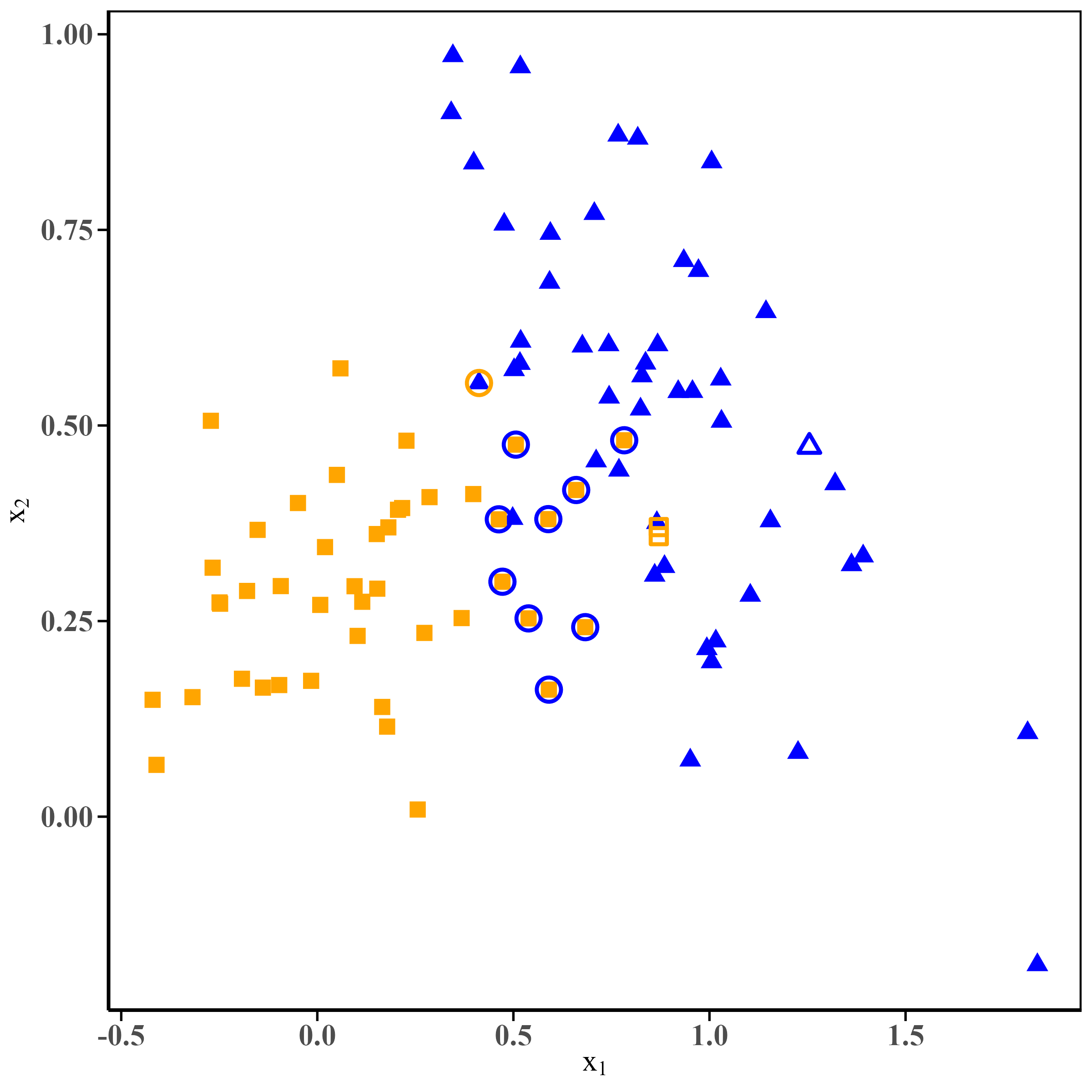}

            \caption{Unified K-MEANS Clustering}
            \label{fig:cluster3}
        \end{subfigure} &
        \begin{subfigure}{0.45\linewidth}
            \centering
            \includegraphics[width=\linewidth]{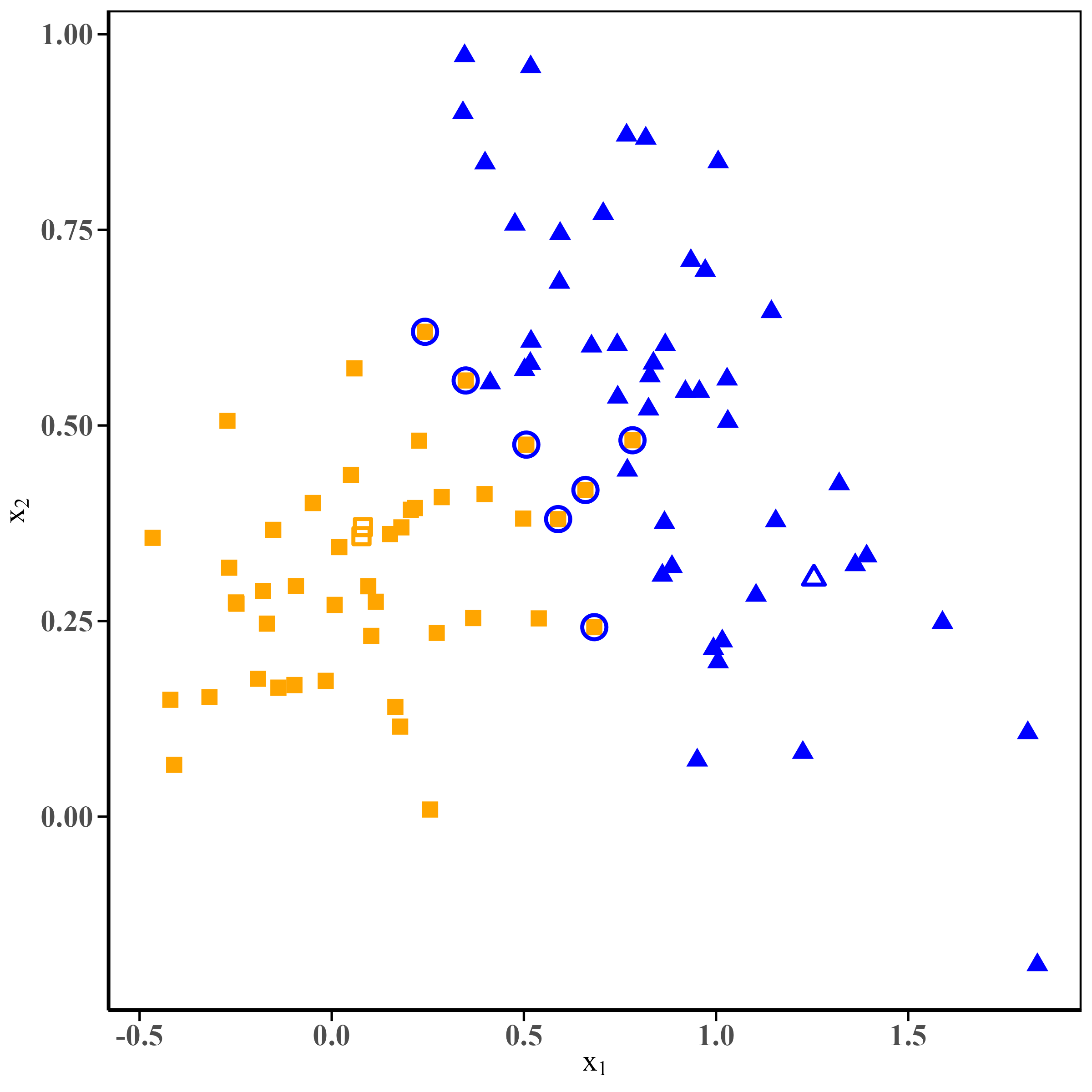}
            \caption{K-MAHAL Clustering}
            \label{fig:cluster4}
        \end{subfigure}
    \end{tabular}
    \caption{The layout of the true cluster distribution (a), along with K-means (b), Unified K-means (c), and K-Mahal (d) using KNN imputation. Class 1 (orange squares) and Class 2 (blue triangles) are generated with $\check{\omega} = 0.1$. Incomplete observations (1.5\%) are indicated by hollow square for Class 1 and hollow triangle for Class 2. Misclassified points are shown with circles.}

    \end{figure}

Figure \ref{fig:cluster1} - \ref{fig:cluster4} shows the Comparison of K-means, Unified K-means, and K-Mahal using KNN imputation for $\check{\omega} = 0.1$ with three incomplete observations (\textit{1.5\%}), indicated by hollow square and hollow triangle. The dataset was randomly generated with \(p=2\) dimensions and \(K=2\) clusters (class 1 (orange squares) and Class 2 (blue triangles)) From (b) (top-right), imputation with K-means resulted in eleven misclassifications (four from the Blue cluster incorrectly classified as Orange and seven from the Orange cluster misclassified as Blue). The bottom-left plot shows that K-means clustering with incomplete data misclassified ten data points (one from the Blue cluster misclassified as Orange and nine from the Orange cluster misclassified as Blue). Additionally, the missing data from the Orange cluster imputed by the algorithm was off the mark. K-Mahal clustering achieved higher clustering accuracy, with a 90\% similarity score compared to the original data. It misclassified seven data points, mostly from the Orange cluster as Blue.

Additionally, the results of clustering experiments on the Iris dataset and a synthetic dataset with \(p=5\) dimensions and \(K=10\) clusters are presented in Tables~\ref{tab:comparison}, \ref{tab:mean_imputation}, \ref{tab:nmi_comparison},  \ref{tab:comparison_generated_knn} and \ref{tab:comparison_generated_mean}. The experiments compare the performance of K-Mahal, Unified K-means, and K-means clustering algorithms across varying percentages of missing data (10\% to 50\%) for one and two incomplete coordinates, different imputation methods (Mean Imputation and KNN Imputation), and various values of the parameter \(\check{\omega}\). Performance is evaluated using Adjusted Rand Index (ARI) and Normalized Mutual Information (NMI). We record the inter-quartile ranges (IQRs) to measure the variability in clustering quality.

\subsection{Clustering Performance with One Coordinate Missing}
Considering the results obtained with Iris dataset using KNN imputation, Table~\ref{tab:nmi_comparison} shows that K-Mahal consistently outperforms Unified K-means and K-means with different levels of data incompleteness. At 10\% missing data, K-Mahal achieves an NMI of 0.914, compared to 0.758 for Unified K-means and 0.758 for K-means. As data incompleteness increases to 50\%, K-Mahal maintains a higher NMI of 0.904, while Unified K-means and K-means decrease to 0.757 and 0.730, respectively.

For the synthetic dataset, Table \ref{tab:comparison_generated_mean} show that K-Mahal achieves an ARI of 0.978 for 10\% missing data when \(\check{\omega}\)=0.001. It significantly outperforms Unified K-means (0.888) and K-means (0.967). Even as the percentage of missing data increases to 50\%, K-Mahal's ARI remains higher at 0.895, demonstrating resilience to missing data. When the overlap increases to $\check{\omega}=0.01$, K-Mahal still outperforms both competitor with ARI = 0.761 even with $50\%$ of incomplete data. 

The IQR values for K-Mahal are lower across all levels of missing data, This indicates consistent clustering performance. 

\subsection{Clustering Performance with Two Coordinates Missing}

When two coordinates are missing, clustering performance declines for all algorithms as the level of incompleteness increases complexity. At 10\% missing data for the Iris dataset, K-Mahal’s NMI drops slightly to 0.901, while Unified K-means and K-means achieve 0.758 and 0.747, respectively as shown by Table \ref{tab:nmi_comparison}. As the percentage of missing data increases, all algorithms degrade significantly with  K-Mahal's NMI  at 0.539 for 50\% missing data, compared to Unified K-means (0.570) and K-means (0.553). Same trends as seen in NMI can also be been for ARI as shown in Tables \ref{tab:mean_imputation} and \ref{tab:comparison}

For the synthetic dataset, with two coordinates missing, cluster performance for all algorithms are poorer compared to one coordinate incomplete as shown in Table \ref{tab:comparison_generated_knn} and \ref{tab:comparison_generated_mean}. For instance, with mean imputation, K-Mahal's ARI drops to 0.940 at 10\% missing data and $\check{\omega}=0.001$ but still outperforms Unified K-means (0.839) and K-means (0.919). As the percentage of missing data increases to 50\%, K-Mahal's ARI further degrades to 0.723, compared to 0.895 for one coordinate missing, yet remains higher than Unified K-means (0.620) and K-means (0.629).

\subsection{Clustering Performance with Imputation Method}

For the generated data set with elliptical clusters, the comparison of the imputation methods (mean and KNN) shows that the imputation of KNN with K-Mahal achieves a consistently higher cluster quality than using the mean imputation, as indicated by the higher values of ARI and NMI at different levels of $\check{\omega}$ and incompleteness of the data.

\subsection{Clustering Performance with Varying Maximum Cluster Separation}

For the generated dataset, different values of "maximum cluster overlap" $\check{\omega}$, were simulated to understand their effect on the performance of K-Mahal, Unified K-means, and K-means clustering algorithms. The results show for instance, at 10\% missing data, K-Mahal achieves an ARI of 0.978, while Unified K-means and K-means record lower values.

However, as $\check{\omega}$ increases to 0.1, which shows a significant overlap between clusters, the KNN imputation proves to be ineffective in higher dimensions, causing the performance gap between the algorithms to narrow and the ARI values drop across the board, although K-Mahal outperforms competitor algorithms

\subsection{Discussion and Extension}
\textbf{Discussion:} The experiments confirm that K-means with Mahalanobis distance, which is known to perform well with elliptical clusters, also works effectively when the data have missing coordinates. Furthermore, our results reinforce the finding that dynamically estimating missing entries during clustering is more effective than separating the imputation and clustering steps.The surprising results of the unified K-means algorithm (K-means with incomplete data as proposed by \cite{wang2019k}), compared to ordinary imputation followed by standard K-means, as also confirmed by Figure~\ref{fig:cluster4}, can be attributed to the fact that the imputed values may have been assigned to a different cluster than their true cluster and then used for clustering. This misalignment likely resulted from ignoring the shape of the clusters. Therefore, with each subsequent iteration, the imputed values deviated further from their true values. This deviation from the true values will negatively impact the performance of Unified K-means.

The degraded performance of our proposed algorithm when dealing with high incompleteness values and in higher-dimensional settings can be attributed to the limitations of KNN and mean imputation as an imputation method in high-dimensional spaces.

\textbf{Extension:} The performance of our method could be further improved by incorporating a specialized imputation technique, which is specifically designed to work with all cluster shapes.

\begin{table*}[ht]
    \centering
    \begin{tabular}{|c|c|c|c|c|c|c|}
        \hline
        \multicolumn{7}{|c|}{\textbf{KNN Imputation}} \\ \hline
        \textbf{Incomplete} & \multicolumn{3}{c|}{\textbf{One Coordinate}} & \multicolumn{3}{c|}{\textbf{Two Coordinates}} \\ \hline
        \textbf{Percentage} & \textbf{K- Mahal} & \textbf{Unified Kmeans} & \textbf{Kmeans} & \textbf{K- Mahal} & \textbf{Unified Kmeans} & \textbf{Kmeans} \\ \hline
        10\%   & 0.922      & 0.73       & 0.73       & 0.922      & 0.742      & 0.717      \\ 
               & [0]        & [0.026]    & [0.014]    & [0.019]    & [0.027]    & [0.014]    \\ \hline
        20\%   & 0.922      & 0.744      & 0.729      & 0.886      & 0.742      & 0.724      \\ 
               & [0.005]    & [0.041]    & [0.014]    & [0.036]    & [0.031]    & [0.038]    \\ \hline
        30\%   & 0.922      & 0.756      & 0.73       & 0.851      & 0.729      & 0.728      \\ 
               & [0.018]    & [0.052]    & [0.027]    & [0.065]    & [0.045]    & [0.047]    \\ \hline
        40\%   & 0.904      & 0.755      & 0.73       & 0.697      & 0.666      & 0.685      \\ 
               & [0.019]    & [0.056]    & [0.027]    & [0.136]    & [0.075]    & [0.075]    \\ \hline
        50\%   & 0.904      & 0.757      & 0.73       & 0.48       & 0.525      & 0.528      \\ 
               & [0.037]    & [0.057]    & [0.028]    & [0.078]    & [0.072]    & [0.076]    \\ \hline
    \end{tabular}
   \caption{Comparison of K-mahal with Unified K-means and K-Means using KNN Imputation with \(d\%\) of observations incomplete in one or two coordinates on the Iris dataset.The table contains the median adjusted Rand Index values with the corresponding IQR shown in brackets.}
    \label{tab:comparison}
\end{table*}

\begin{table*}[ht]
    \centering
    \begin{tabular}{|c|c|c|c|c|c|c|}
        \hline
        \multicolumn{7}{|c|}{\textbf{Mean Imputation}} \\ \hline
        \textbf{Incomplete} & \multicolumn{3}{c|}{\textbf{One Coordinate}} & \multicolumn{3}{c|}{\textbf{Two Coordinates}} \\ \hline
        \textbf{Percentage} & \textbf{K- Mahal} & \textbf{Unified Kmeans} & \textbf{Kmeans} & \textbf{K- Mahal} & \textbf{Unified Kmeans} & \textbf{Kmeans} \\ \hline
        \multirow{2}{*}{10\%}  & 0.903  & 0.730  & 0.694  & 0.847  & 0.707  & 0.673  \\ 
                                & [0.042] & [0.028] & [0.035] & [0.038] & [0.047] & [0.043] \\ \hline
        \multirow{2}{*}{20\%}  & 0.846  & 0.729  & 0.694  & 0.739  & 0.667  & 0.642  \\ 
                                & [0.060] & [0.034] & [0.057] & [0.093] & [0.052] & [0.076] \\ \hline
        \multirow{2}{*}{30\%}  & 0.761  & 0.710  & 0.660  & 0.634  & 0.645  & 0.583  \\ 
                                & [0.096] & [0.048] & [0.070] & [0.099] & [0.076] & [0.081] \\ \hline
        \multirow{2}{*}{40\%}  & 0.734  & 0.704  & 0.643  & 0.333  & 0.496  & 0.319  \\ 
                                & [0.120] & [0.060] & [0.066] & [0.066] & [0.090] & [0.057] \\ \hline
        \multirow{2}{*}{50\%}  & 0.667  & 0.686  & 0.612  & 0.192  & 0.367  & 0.165  \\ 
                                & [0.112] & [0.068] & [0.069] & [0.094] & [0.089] & [0.044] \\ \hline
    \end{tabular}
    \caption{Comparison of K-Mahal with Unified K-means and K-Means using Mean Imputation with \(d\%\) of observation missing in one or two coordinates on Iris dataset. The table contains the median adjusted Rand Index values with the corresponding IQR shown in brackets.}
    \label{tab:mean_imputation}
\end{table*}

\begin{table*}[ht]

\begin{tabular}{|c|c|c|c|c|c|c|}
\hline
\textbf{Incomplete \%} & \multicolumn{3}{c|}{\textbf{One coordinate}} & \multicolumn{3}{c|}{\textbf{Two coordinates}} \\ \hline
& \textbf{K-Mahal} & \textbf{Unified Kmeans} & \textbf{Kmeans} & \textbf{K-Mahal} & \textbf{Unified Kmeans} & \textbf{Kmeans} \\ \hline
10\% & 0.914 & 0.758 & 0.758 & 0.901 & 0.758 & 0.747 \\ 
   & {[0]} & {[0.008]} & {[0.016]} & {[0.029]} & {[0.017]} & {[0.016]} \\ \hline
20\% & 0.914 & 0.762 & 0.751 & 0.874 & 0.758 & 0.744 \\ 
   & {[0.015]} & {[0.02]} & {[0.016]} & {[0.042]} & {[0.029]} & {[0.025]} \\ \hline
30\% & 0.914 & 0.767 & 0.758 & 0.841 & 0.748 & 0.742 \\ 
   & {[0.029]} & {[0.025]} & {[0.019]} & {[0.052]} & {[0.031]} & {[0.038]} \\ \hline
40\% & 0.900 & 0.770 & 0.758 & 0.697 & 0.691 & 0.683 \\ 
   & {[0.034]} & {[0.03]} & {[0.016]} & {[0.094]} & {[0.057]} & {[0.057]} \\ \hline
50\% & 0.904 & 0.757 & 0.730 & 0.539 & 0.570 & 0.553 \\ 
   & {[0.037]} & {[0.057]} & {[0.028]} & {[0.055]} & {[0.058]} & {[0.067]} \\ \hline
\end{tabular}
\centering
\caption{Comparison of K-mahal with Unified K-means and K-Means using KNN Imputation with \(d\%\) of observation incomplete in one or two coordinates on the Iris dataset.The table contains the median Normalized Mutual Index (NMI) with the corresponding IQRs shown in bracket}
\label{tab:nmi_comparison}
\end{table*}

%\begin{figure}[hbt]
%    \centering
%    \includegraphics [width=1.0\textwidth]{NMIplot.pdf}
%     \caption{ Median Normalized Mutual Index (NMI) Trends with \(d\%\) of observation incomplete in one or two coordinates on the Iris dataset. }
 %   \label{maxFraction}
%\end{figure}

\begin{table*}[ht]
    \centering
    \begin{tabular}{|c|ccc|ccc|}
        \hline
        \multirow{2}{*}{\textbf{$\check{\omega}, d\%$}} & \multicolumn{3}{c|}{\textbf{One Coordinate}} & \multicolumn{3}{c|}{\textbf{Two Coordinates}} \\ \cline{2-7} 
                                             & \textbf{K-Mahal} & \textbf{Unified K-means} & \textbf{K-means} & \textbf{K-Mahal} & \textbf{Unified K-means} & \textbf{K-means} \\ \hline
        0.001,10\%   & 1.000     & 0.876     & 0.883     & 0.986     & 0.867     & 0.974     \\ 
               & [0.002]  & [0.017]  & [0.120]   & [0.007]  & [0.112]  & [0.118]  \\ \hline
        0.001,20\%   & 0.997    & 0.911    & 0.933     & 0.951    & 0.828    & 0.950    \\ 
               & [0.002]  & [0.117]  & [0.127]   & [0.015]  & [0.077]  & [0.048]  \\ \hline
        0.001,30\%   & 0.996    & 0.897    & 0.936     & 0.898    & 0.803    & 0.898    \\ 
               & [0.003]  & [0.109]  & [0.120]   & [0.024]  & [0.089]  & [0.021]  \\ \hline
        0.001,40\%   & 0.993    & 0.897    & 0.932     & 0.854    & 0.760    & 0.854    \\ 
               & [0.004]  & [0.111]  & [0.119]   & [0.029]  & [0.049]  & [0.025]  \\ \hline
        0.001,50\%   & 0.992    & 0.890    & 0.912     & 0.804    & 0.701    & 0.805    \\ 
               & [0.008]  & [0.100]  & [0.125]   & [0.025]  & [0.054]  & [0.027]  \\ \hline
        \hline
        0.01, 10\%     & 0.994    & 0.983    & 0.889    & 0.976    & 0.852    & 0.875   \\ 
               & [0.002]  & [0.075]  & [0.088]   & [0.014]  & [0.096]  & [0.103]  \\ \hline
        0.01,20\%   & 0.992    & 0.970    & 0.934     & 0.949    & 0.839    & 0.868    \\ 
               & [0.002]  & [0.034]  & [0.114]   & [0.080]  & [0.084]  & [0.095]  \\ \hline
        0.01,30\%   & 0.992    & 0.948    & 0.885     & 0.906    & 0.814    & 0.893    \\ 
               & [0.005]  & [0.094]  & [0.085]   & [0.026]  & [0.075]  & [0.071]  \\ \hline
        0.01,40\%   & 0.986    & 0.891    & 0.879     & 0.832    & 0.728    & 0.815    \\ 
               & [0.005]  & [0.091]  & [0.074]   & [0.040]  & [0.018]  & [0.062]  \\ \hline
        0.01,50\%   & 0.982    & 0.937    & 0.881     & 0.784    & 0.685    & 0.770    \\ 
               & [0.008]  & [0.085]  & [0.103]   & [0.041]  & [0.052]  & [0.061]  \\ \hline
        \hline
        0.1, 10\%     & 0.857    & 0.793    & 0.844    & 0.818    & 0.795    & 0.814   \\ 
               & [0.008]  & [0.056]  & [0.009]   & [0.020]  & [0.020]  & [0.012]  \\ \hline
        0.1,20\%   & 0.850    & 0.812    & 0.829     & 0.768    & 0.699    & 0.748    \\ 
               & [0.003]  & [0.025]  & [0.047]   & [0.014]  & [0.037]  & [0.024]  \\ \hline
        0.1,30\%   & 0.837    & 0.762    & 0.817     & 0.704    & 0.627    & 0.696    \\ 
               & [0.013]  & [0.029]  & [0.045]   & [0.030]  & [0.039]  & [0.030]  \\ \hline
        0.1,40\%   & 0.814    & 0.743    & 0.778     & 0.655    & 0.581    & 0.636    \\ 
               & [0.025]  & [0.048]  & [0.042]   & [0.014]  & [0.023]  & [0.019]  \\ \hline
        0.1,50\%   & 0.808    & 0.696    & 0.785     & 0.588    & 0.506    & 0.573    \\ 
               & [0.003]  & [0.018]  & [0.014]   & [0.043]  & [0.033]  & [0.037]  \\ \hline
    \end{tabular}
    \caption{Comparison of K-Mahal with Unified K-means and K-means using KNN Imputation and with \(d\%\) of observations incomplete in one or two coordinate(s) and different values of pairwise overlap. The dataset is generated randomly with \(p=5\) dimensions, \(K=10\) clusters. The table contains the median adjusted Rand Index values with the corresponding IQR shown in brackets.}
    \label{tab:comparison_generated_knn}
\end{table*}

\begin{table*}[ht]
    \centering
    \begin{tabular}{|c|c|ccc|ccc|}
        \hline
        \multirow{2}{*}{$\check{\omega}$} & \multirow{2}{*}{\textbf{Incomplete \%}} & \multicolumn{3}{c|}{\textbf{One Coordinate}} & \multicolumn{3}{c|}{\textbf{Two Coordinates}} \\ 
        \cline{3-8} 
        &  & \textbf{K-Mahal} & \textbf{Unified K-means} & \textbf{K-means} & \textbf{K-Mahal} & \textbf{Unified K-means} & \textbf{K-means} \\ \hline

        \multirow{5}{*}{0.001} 
        & 10\%  & 0.978 & 0.888 & 0.967 & 0.940 & 0.839 & 0.919 \\ 
        &       & [0.009] & [0.105] & [0.107] & [0.015] & [0.096] & [0.023] \\ 
        & 20\%  & 0.955 & 0.875 & 0.925 & 0.884 & 0.798 & 0.842 \\ 
        &       & [0.010] & [0.086] & [0.028] & [0.029] & [0.047] & [0.035] \\ 
        & 30\%  & 0.939 & 0.874 & 0.913 & 0.823 & 0.736 & 0.767 \\ 
        &       & [0.014] & [0.089] & [0.025] & [0.023] & [0.045] & [0.023] \\ 
        & 40\%  & 0.913 & 0.844 & 0.881 & 0.774 & 0.670 & 0.696 \\ 
        &       & [0.016] & [0.078] & [0.035] & [0.016] & [0.064] & [0.019] \\ 
        & 50\%  & 0.895 & 0.808 & 0.866 & 0.723 & 0.620 & 0.629 \\ 
        &       & [0.028] & [0.069] & [0.024] & [0.028] & [0.047] & [0.034] \\ \hline

        \multirow{5}{*}{0.01} 
        & 10\%  & 0.928 & 0.861 & 0.960 & 0.888 & 0.837 & 0.914 \\ 
        &       & [0.081] & [0.025] & [0.089] & [0.068] & [0.062] & [0.069] \\ 
        & 20\%  & 0.941 & 0.885 & 0.916 & 0.873 & 0.847 & 0.870 \\ 
        &       & [0.010] & [0.118] & [0.070] & [0.017] & [0.083] & [0.082] \\ 
        & 30\%  & 0.926 & 0.850 & 0.940 & 0.809 & 0.733 & 0.781 \\ 
        &       & [0.016] & [0.080] & [0.065] & [0.005] & [0.029] & [0.028] \\ 
        & 40\%  & 0.903 & 0.853 & 0.896 & 0.752 & 0.661 & 0.699 \\ 
        &       & [0.025] & [0.040] & [0.065] & [0.005] & [0.030] & [0.047] \\ 
        & 50\%  & 0.877 & 0.800 & 0.883 & 0.675 & 0.597 & 0.646 \\ 
        &       & [0.008] & [0.080] & [0.038] & [0.000] & [0.042] & [0.041] \\ \hline

        \multirow{5}{*}{0.1} 
        & 10\%  & 0.846 & 0.819 & 0.833 & 0.807 & 0.779 & 0.787 \\ 
        &       & [0.006] & [0.050] & [0.008] & [0.015] & [0.008] & [0.030] \\ 
        & 20\%  & 0.822 & 0.777 & 0.794 & 0.741 & 0.683 & 0.694 \\ 
        &       & [0.012] & [0.029] & [0.020] & [0.010] & [0.026] & [0.028] \\ 
        & 30\%  & 0.807 & 0.744 & 0.779 & 0.684 & 0.617 & 0.652 \\ 
        &       & [0.016] & [0.044] & [0.023] & [0.011] & [0.032] & [0.024] \\ 
        & 40\%  & 0.782 & 0.682 & 0.724 & 0.632 & 0.531 & 0.584 \\ 
        &       & [0.012] & [0.042] & [0.059] & [0.036] & [0.020] & [0.021] \\ 
        & 50\%  & 0.761 & 0.662 & 0.729 & 0.584 & 0.499 & 0.507 \\ 
        &       & [0.010] & [0.015] & [0.015] & [0.041] & [0.051] & [0.030] \\ \hline
    \end{tabular}
    \caption{Comparison of K-Mahal with Unified K-means and K-means using Mean Imputation for varying $\check{\omega}$ and different percentages of incomplete observations (\textit{d\%}). The dataset was randomly generated with \(p=5\) dimensions and \(K=10\) clusters. The table contains the median adjusted Rand Index values with the corresponding interquartile range (IQR) shown in brackets.}
    \label{tab:comparison_generated_mean}
\end{table*}

%\begin{figure}[hbt]
   % \centering
   % \includegraphics [width=1.0\textwidth]{KmeanPic.pdf}
    %  \caption{Adjusted Rand Index (ARI) trends for different values of pairwise overlap with one and two coordinates incomplete for  K-means  using Mean Imputation.}
   % \label{fig:ari_trends_separate_kmeanPic}
%\end{figure}

%\begin{figure}[hbt]
   % \centering
   % \includegraphics [width=1.0\textwidth]{UnifiedPic.pdf}
    % \caption{Adjusted Rand Index (ARI) trends for different values of pairwise overlap with one and two coordinates incomplete for Unified K-means using Mean Imputation.}
   % \label{fig:ari_trends_separate_Unified}
%\end{figure}

%\begin{figure}[hbt]
 %   \centering
 %   \includegraphics [width=1.0\textwidth]{K_Mahal_Plot.pdf}
 %     \caption{Adjusted Rand Index (ARI) trends for different values of pairwise overlap with one and two coordinates incomplete for K-Mahal  using Mean Imputation.}
 %   \label{fig:ari_trends_separate_Kmahal}
%\end{figure}

\subsection{Conclusion}
\label{sec:conclusion}
We demonstrated that for elliptical clusters with incomplete data, K-means clustering with incomplete data using Mahalanobis distance should be preferred. In the future, we aim to develop a specialized imputation technique that is tailored for elliptical clusters, which will further improve the algorithm's performance when dealing with a higher number of missing coordinates.

\bibliographystyle{plain}
\bibliography{references}

\end{document}